\documentclass[sigconf]{acmart}
\setcopyright{none}
\renewcommand\footnotetextcopyrightpermission[1]{}

\usepackage{multirow}
\usepackage[caption=true, font=footnotesize]{subfig}
\usepackage{rotating}
\usepackage{xcolor}
\usepackage{todonotes}

\AtBeginDocument{%
  \providecommand\BibTeX{{%
    \normalfont B\kern-0.5em{\scshape i\kern-0.25em b}\kern-0.8em\TeX}}}

\newcommand{\guanxiong}[1]{\textcolor{black}{#1}}       
\newcommand{\guanxiongA}[1]{\textcolor{black}{#1}}       

\begin{document}

\title{Using Single-Step Adversarial Training to Defend Iterative Adversarial Examples}

\author{Guanxiong Liu}
\affiliation{%
  \institution{New Jersey Institute of Technology}
  \streetaddress{University Heights}
  \city{Newark}
  \state{NJ}
  \postcode{07102}
}
\email{gl236@njit.edu}

\author{Issa Khalil}
\affiliation{%
  \institution{Qatar Computing Research Institute}
  \city{Doha}
  \state{Qatar}
}
\email{ikhalil@hbku.edu.qa}

\author{Abdallah Khreishah}
\affiliation{%
  \institution{New Jersey Institute of Technology}
  \streetaddress{University Heights}
  \city{Newark}
  \state{NJ}
  \postcode{07102}
}
\email{abdallah@njit.edu}


\begin{abstract}
Adversarial examples have become one of the largest challenges that machine learning models, especially neural network classifiers, face. These adversarial examples break the assumption of attack-free scenario and fool state-of-the-art (SOTA) classifiers with insignificant perturbations to human. So far, researchers achieved great progress in utilizing adversarial training as a defense. However, the overwhelming computational cost degrades its applicability and little has been done to overcome this issue. Single-Step adversarial training methods have been proposed as computationally viable solutions, however they still fail to defend against iterative adversarial examples. In this work, we first experimentally analyze several different SOTA defense methods against adversarial examples. Then, based on observations from experiments, we propose a novel single-step adversarial training method which can defend against both single-step and iterative adversarial examples. \guanxiongA{Lastly, through extensive evaluations, we demonstrate that our proposed method outperforms the SOTA \textbf{single-step} and \textbf{iterative} adversarial training defense. Compared with ATDA (single-step method) on CIFAR10 dataset, our proposed method achieves 35.67\% enhancement in test accuracy and 19.14\% reduction in training time. When compared with methods that use BIM or Madry examples (iterative methods) on CIFAR10 dataset, it saves up to 76.03\% in training time with less than 3.78\% degeneration in test accuracy.}
\end{abstract}
\pagestyle{plain}



\maketitle

\section{Introduction}\label{sec:introduction}

Adversarial examples were discovered and presented in \cite{szegedy2013intriguing} under image classification tasks by showing that specially designed insignificant perturbations can effectively mislead neural network (\textbf{NN}) classifiers. More importantly, it has been shown that such perturbations are not special cases but rather generic and can be crafted for almost every input example. Yet, more scary, researchers show that adversaries could arbitrarily control the prediction results as desired with high success rate through carefully designed perturbations \cite{carlini2017towards}\cite{kurakin2016adversarial1}.


\guanxiong{
Thereafter, great effort has been devoted to designing defense methods against adversarial examples. Some of these methods utilize augmentation and regularization to enhance test accuracy on adversarial examples \cite{papernot2016distillation}. Other methods intercepts the input to the classifier to either identify and drop adversarial examples or eliminate adversarial perturbations \cite{meng2017magnet}\cite{samangouei2018defense}. Currently, the most popular choices of defense methods are based on adversarial training \cite{madry2017towards}.
}

\guanxiong{
The fundamental idea of adversarial training is to use adversarial examples as blind spots and retrain the classifier with them. This enhances the capability of the classifier to correctly classify the perturbed examples to their original labels. These defenses could be categorized based on the adversarial examples they are using. Defenses that use \textbf{single-step} adversarial examples are called \textbf{Single-Adv}, while defenses that use \textbf{iterative} adversarial examples called \textbf{Iter-Adv}.
}

\guanxiong{
Among all existing defense approaches, adversarial training is shown to be the most successful solution, because unlike many others, it does not rely on the false sense of security brought by obfuscated gradient \cite{athalye2018obfuscated}. However, adversarial training still has many unsolved problems. The high computational cost in preparing adversarial examples during training \cite{liu2019zk} is among the most challenging ones. If Single-Adv defense is used, the trained model will not be able to defend against iterative adversarial examples \cite{kurakin2016adversarial1}. On the other hand, Iter-Adv defenses can defend against iterative examples, however, they require powerful GPU infrastructure for an ImageNet like datasets \cite{kannan2018adversarial}. 
}

\guanxiong{
Nowadays, more and more smartphone and smart-home applications are integrated with machine learning components, especially the popular NN classifier. These applications are running in an environment without sufficient local computing power for adversarial training. Without considering the cost, a cloud server could partially solve this issue. However, some of these applications, that are running with sensitive or personal data, are not willing to share their data and model with a cloud server. Therefore, these applications lack the tools to defend against adversarial examples.}

In general, current adversarial training methods fail to achieve robust performance with acceptable training overhead. In the following sections, we first provide detailed analysis of the state-of-the-art (\textbf{SOTA}) defense methods. \guanxiongA{Then, based on lessons learned from analysis, we propose a Single-Adv method that can efficiently mitigate iterative adversarial example with low training overhead and name it \textbf{S}ingle-Step Epoch-\textbf{I}terative \textbf{M}ethod (\textbf{SIM-Adv}). Intuitively, it flattens iterative adversarial examples (\textbf{Iter-Exps}) into single-step adversarial examples (\textbf{Single-Exps}) in multiple consecutive training epochs as detailed later.} Our contributions are summarized as follows:


\begin{itemize}
    \item We analyze the SOTA Single-Adv method called ATDA. Our results indicate that the evaluation of ATDA as presented in \cite{song2018improving} was incomplete because we show that it fails to defend against many adversarial examples.
    
    \item \guanxiongA{We analyze the SOTA Iter-Adv methods and identify guidelines that can reduce computational overhead while achieving good performance in defending against Iter-Exps. The guidelines include: (1) using large per step perturbation and (2) utilizing the intermediate examples during the preparation of Iter-Exps.}
    
    \item Based on the observations from the analysis, we propose a novel Single-Adv method which is the first in this category that can defend against both Single-Exps and Iter-Exps while maintaining very low training overhead. This method opens the door for the applications with limited computing power to build computationally efficient self defenses against adversarial examples. 
\end{itemize}

The rest of the work is organized as follows. Section \ref{sec:background} summarizes important background material. Sections \ref{sec:atda} and \ref{sec:iterative} present detailed analysis of different SOTA adversarial training methods. Section \ref{sec:methodology} introduces our cost-effective Single-Adv method. Section \ref{sec:evaluation} presents evaluation results and Section \ref{sec:conclusion} concludes the paper.

\section{Background}\label{sec:background}

In this section, we review the fundamental material and provide references for further understanding of the concepts presented in this work.

\subsection{Adversarial Examples}

As mentioned in Section \ref{sec:introduction}, adversarial examples are specially perturbed original examples which aim to fool the NN classifier. Although adversarial examples can be categorized according to different aspects, we focus here on the distinction between Single-Exps and Iter-Exps. Adversarial examples throughout this work are $l_{\infty}$ white-box untargeted adversarial examples in the image classification domain. In other words: (1) the perturbation of adversarial example is limited in an $l_{\infty}$ norm ball, (2) the classifier is transparent to adversary, and (3) the adversary successfully fools the classifier as long as the prediction is wrong.

\subsubsection{Single-Step Adversarial Examples:}

In the early stage of adversarial example research, \cite{goodfellow2014explaining} introduces the Fast Gradient Sign Method (\textbf{FGSM}) to generate Single-Exps. FGSM calculates gradients of the classifier's loss function, $L$, towards each original example, $\hat{x}$. Then, it takes the sign of the calculated gradients and multiplies it with the perturbation limit, $\epsilon$. This product is called adversarial perturbation, $\delta$, and adding it to the original example can turn it into adversarial example, $\tilde{x}$. The mathematical formulation of FGSM is as follows.
\begin{align}
    & \delta = clip_{[-\epsilon, \epsilon]} [\epsilon \times sign[\nabla_{\hat{x}} L(\hat{x}, y, \theta)]] \\
    & \tilde{x} = clip_{[0,1]} [\hat{x} + \delta]
\end{align}
Here, $y$ is the ground truth label, $\theta$ is the parameter of the neural network classifier, and $clip$ is a function that maps out-of-range values to the closest boundary and preserves the in-range values.

As can be seen from the formulation, FGSM utilizes a linear approximation of the loss function value which only calculates gradients once during the generation process. Therefore, it is a cost-efficient method to generate adversarial examples and is widely used in works like \cite{kurakin2016adversarial1} and \cite{madry2017towards}.

In addition to FGSM, there are many other methods proposed to generate Single-Exps. For example, researchers in \cite{tramer2017ensemble} introduce the Random Initialized FGSM (\textbf{R+FGSM}) as an attack method to break defenses that rely on the gradient masking effect.

\subsubsection{Iterative Adversarial Examples:}

Compared with Single-Exps, Iter-Exps are more serious adversarial examples since they are harder to be mitigated. Instead of linear approximation on the loss function value, Iter-Exps are generated by smaller per step perturbations ($\frac{\epsilon}{n_{1}}$) of the original examples based on gradient calculations of the loss function , where $n_{1}$ is a scale factor. The small perturbation process is repeated for a certain number of iterations, $n_{2}$. Two widely used methods for generating Iter-Exps are the Basic Iterative Method (\textbf{BIM}) and the Madry Method (\textbf{Madry}) which are introduced in \cite{kurakin2016adversarial1} and \cite{madry2017towards}, respectively. These two methods utilize the projected gradient descent method to perturb the example during each iteration. The only difference between BIM and Madry is that Madry randomly perturbs the original example and utilizes it as the starting point, $\tilde{x}_{0}$, while BIM always starts with the original example itself. The mathematical formulations are listed below.
\begin{align}
    & \delta_{i+1} = clip_{[-\epsilon,\epsilon]} [\frac{\epsilon}{n_{1}} \times sign[\nabla_{\tilde{x}_{i}} L(\tilde{x}_{i}, y, \theta)]] ~~ i \in [0,n_{2}] \\
    & \tilde{x}_{i+1} = clip_{[0,1]} [\tilde{x}_{i} + \delta_{i+1}] ~~ \tilde{x}_{0} = \hat{x} ~~ \tilde{x} = \tilde{x}_{n_{2}}  \label{eq:iter-exps}
\end{align}

Throughout this work, we represent the different Iter-Exps generation methods in the following format, ``Name($n_{1}, n_{2}$)''. For example, BIM with $n_{1} = n_{2} = 10$ will be represented as BIM(10,10).

\subsection{Adversarial Training}

The fundamental idea of adversarial training is to use adversarial examples as blind spots and retrain the classifier with these blind spots. From a system level point of view, adversarial training can be represented as a two-step process.
\begin{align}
    & \delta_{i+1} = \arg \max_{\delta \in \Delta} I[C(x + \delta | \theta_{i}) \neq y] \label{eq:find-delta} \\
    & \theta_{i+1} = \arg \min_{\theta} L(x + \delta_{i+1}, y, \theta) \label{eq:find-theta}
\end{align}

Here, $I$ is an indicator function, $C$ represents the classifier, $L$ is a loss function, and $\Delta$ is the feasible set of $\delta$. Starting with a randomly initialized $\theta_{0}$, the searching of adversarial perturbation (Eq. \ref{eq:find-delta}) and the training of classifier (Eq. \ref{eq:find-theta}) will be competing with each other repeatedly until $\theta$ converges.

\subsubsection{Single-step Adversarial Training:}
Early adversarial training research retrain with Single-Exps (\textbf{Single-Adv}). For example in \cite{goodfellow2014explaining}, researchers apply adversarial training with FGSM examples (\textbf{FGSM-Adv}), while the work in \cite{tramer2017ensemble} enhances the FGSM-Adv through generating FGSM examples based on holdout classifiers. However, these two as well as many other Single-Adv methods fail to defend against Iter-Exps. Recently, a SOTA Single-Adv method (\textbf{ATDA}) is introduced in \cite{song2018improving} which claims to defend against both Single-Exps and Iter-Exps. To achieve this, the method utilizes FGSM examples and a domain adaptation loss function. However, as we show later, the domain adaptation does not help adversarial training and hence the ATDA trained classifier fails to defend against Iter-Exps.

\subsubsection{Iterative Adversarial Training:}
To defend against both Single-Exps and Iter-Exps, researchers combine adversarial training with Iter-Exps (\textbf{Iter-Adv}). Among these methods, adversarial training with BIM examples (BIM-Adv) and Madry examples (Madry-Adv) are common choices due to their relatively high classification accuracy as shown in \cite{kurakin2016adversarial1} and \cite{madry2017towards}, respectively. Based on these two defense methods, the work in \cite{kannan2018adversarial} claims to improve the defense by utilizing a logit pairing loss function. However, as mentioned in \cite{athalye2018obfuscated} and \cite{kurakin2016adversarial2}, all these Iter-Adv methods are computationally expensive. This problem makes Iter-Adv methods not very practical, especially in platforms with limited computing power.

\section{Analysis of the ATDA Method}\label{sec:atda}

Single-Adv methods have the advantage of less computation overhead compared with Iter-Adv methods. However, most of Single-Adv methods fail to defend against Iter-Exps. However, a recent method dubbed ATDA has been introduced in \cite{song2018improving} as the SOTA Single-Adv method that defends against both Single-Exps and Iter-Exps by adding the so-called domain adaptation loss. \guanxiongA{In the following, we thoroughly analyze ATDA by presenting its detailed prediction on Iter-Exps (BIM and Madry examples) and comparing with other defenses.}

\subsection{Details of the ATDA Method}

\guanxiong{
The fundamental idea of ATDA is to combine adversarial training with domain adaptation. The authors of ATDA believe that Single-Exps (e.g. FGSM examples) could be envisioned as limited number of samples in the adversarial domain. Therefore, combining domain adaptation methods could enhance the performance of Single-Adv. In its design, ATDA utilizes both unsupervised and supervised domain adaptations as follows.
}

\subsubsection{Unsupervised Domain Adaptation:}
\guanxiong{
The authors assume that both the predictions of original examples, $\hat{X} = \{\hat{x}\}$, and adversarial examples, $\tilde{X} = \{\tilde{x}\}$, follow two different multivariate normal distributions. To align these two distributions together, the authors defined a covariance distance as
\begin{align}
    & \mathcal{L}_{CORAL} = \frac{1}{k^{2}} || Cov[C(\hat{X})] - Cov[C(\tilde{X})] ||_{l_{1}}
\end{align}
where $C(\cdot)$ represents the predictions, $Cov[\cdot]$ represents the covariance matrix and $k$ denotes the number of classes in the predictions. In addition to that, authors also utilize the standard distribution distance metric in \cite{borgwardt2006integrating} and the sum of these two distances is defined as the unsupervised domain adaptation loss.
\begin{align}
    & \mathcal{L}_{MMD} = \frac{1}{k} || \frac{1}{|\hat{X}|} \underset{\hat{x} \in \hat{X}}{\Sigma} C(\hat{x}) - \frac{1}{|\tilde{X}|} \underset{\tilde{x} \in \tilde{X}}{\Sigma} C(\tilde{x}) ||_{1} \\
    & \mathcal{L}_{UDA} = \mathcal{L}_{CORAL} + \mathcal{L}_{MMD}
\end{align}
}

\subsubsection{Supervised Domain Adaptation:}
\guanxiong{
In order to utilize the ground truth information in the training dataset, ATDA proposes a new loss function to minimize the intra-class variations and maximize the inter-class variations.
\begin{align}
    \mathcal{L}_{SDA} &= \frac{1}{(k-1) (|\hat{X}| + |\tilde{X}|)} \times \\
    & \underset{x \in \hat{X} \cup \tilde{X}}{\Sigma} \underset{\phi_{m} \in \Phi \setminus \{\phi_{x}\}}{\Sigma} softplus(||C(x)-\phi_{x}||_{1} - ||C(x)-\phi_{m}||_{1})
\end{align}
The $|\cdot|$ counts the number of examples in a set. The $softplus$ represents the function $\ln(1 + \exp(\cdot))$. The $\phi_{x}$ is the center of predictions from examples that has the same ground truth as example $x$. The centers of predictions from examples in each ground truth class are collected in a set which is denoted by $\Phi = \{\phi_{m}|m=1,2,...,k\}$. These centers are also updated together with parameters in the NN based on the following equation.
\begin{align}
    & \phi_{m} \leftarrow \phi_{m} - \beta \cdot \frac{\underset{x \in (\hat{X} \cup \tilde{X})_{y = m}}{\Sigma} (\phi_{m} - C(x)) }{1 + | (\hat{X} \cup \tilde{X})_{y = m} |}
\end{align}
Here, $(\hat{X} \cup \tilde{X})_{y = m}$ represents the subset of original and adversarial examples that belongs to $m$th class. Also, $\beta$ denotes the step size of the update and is set to be 0.1 according to \cite{song2018improving}.
}

\begin{figure*}[ht]
\centering
\begin{minipage}[c]{.95\textwidth}
    \begin{minipage}[c]{\textwidth}
    \centering
        \includegraphics[width=\linewidth]{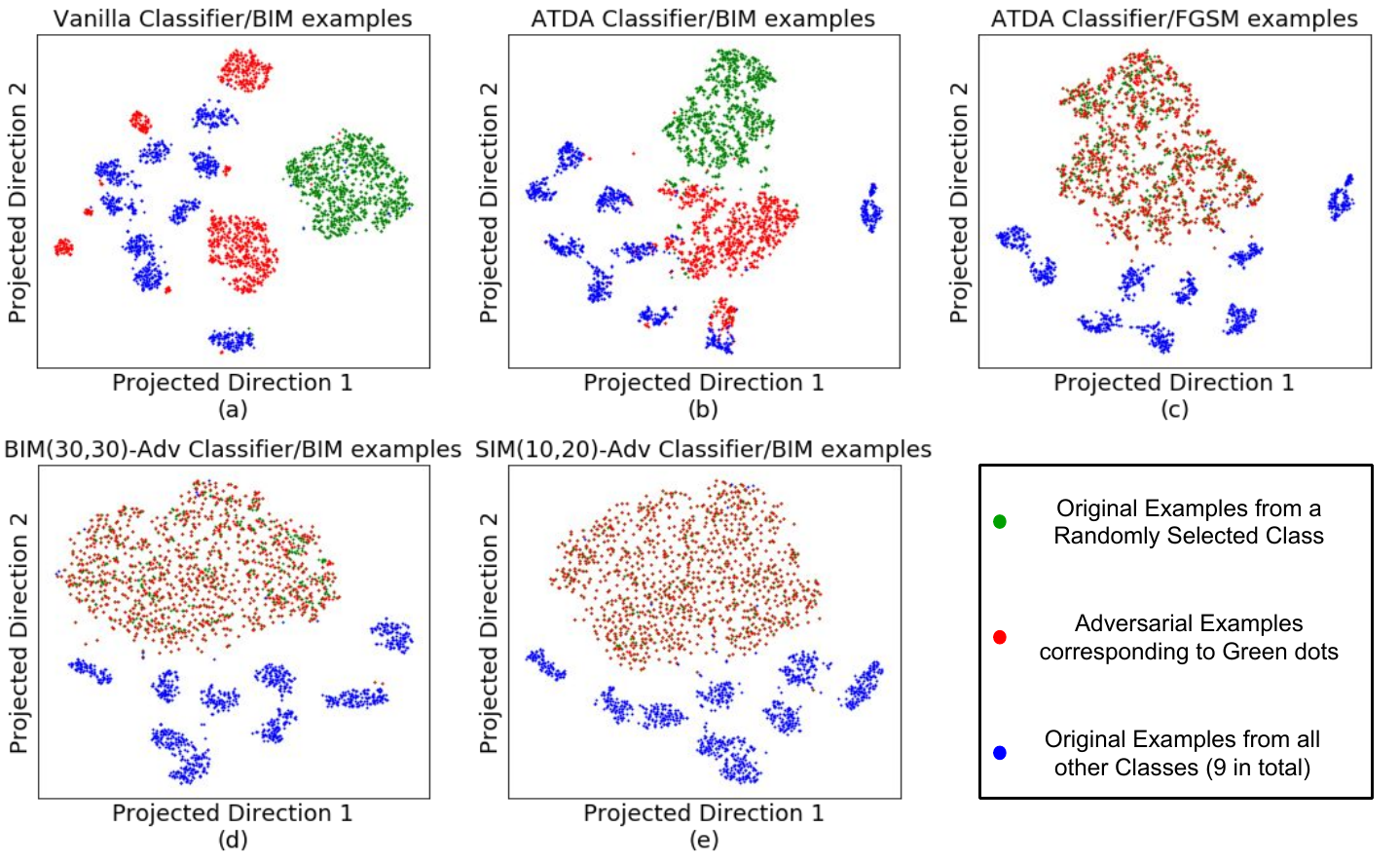}
    \end{minipage}
\caption{Comparing the Feature Space Encoding of Vanilla, ATDA, BIM(30,30)-Adv, and SIM(10,20)-Adv Classifiers}
\label{fig:atda-toy}
\end{minipage}
\end{figure*}
\subsection{Analyzing Feature Space Encoding}

The first step of our analysis targets the feature space encoding from four different classifiers (vanilla, ATDA, BIM(30,30)-Adv, and SIM(10,20)-Adv classifiers). We use t-SNE \cite{maaten2008visualizing} to project the high dimensional feature encoding from each classifier to a two dimensional space and visualize it in Figure \ref{fig:atda-toy}. During the analysis, we sample original examples from MNIST dataset. Examples from a randomly selected class are used as targets (green dots) while others are references (blue dots). Corresponding to targets, we generate adversarial examples (red dots). In the visualized feature space encoding, examples that are close to each other and form a group are much likely to be classified in the same class. Therefore, the classifier that can defend against adversarial examples should have a visualization where green and red dots are blended together.

From Figure~\ref{fig:atda-toy}a, we can see that the feature encoding of adversarial examples (BIM(30,30) examples) form several individual small groups are clearly different from the feature encoding of targets (green dots). Without the color, we can barely tell the difference between small groups of references (blue dots) and adversarial examples (red dots). When using the ATDA classifier, this problem is slightly mitigated since the red dots are grouped together in Figure \ref{fig:atda-toy}b. However, from Figure \ref{fig:atda-toy}b, the groups of green dots and red dots are still separable. BIM(30,30)-Adv and our SIM(10,20)-Adv classifiers presented in Figures \ref{fig:atda-toy}d and \ref{fig:atda-toy}e are significantly better than the ATDA classifier since the red and green dots are blended. It's worth recalling that our SIM(10,20)-Adv belongs to Single-Adv and takes less training time than ATDA as presented later. Therefore, our method outperforms ATDA in both accuracy and training time as detailed in later sections. Finally, we add an extra visualization for the ATDA classifier and change the adversarial examples to FGSM examples. Compared with BIM(30,30) examples which are Iter-Exps, the FGSM examples are weaker Single-Exps. In Figure \ref{fig:atda-toy}c, the red and green dots are blended which means, as expected, that the ATDA classifier performs well against FGSM (Single-Exps) examples.

\subsection{Overfitting to FGSM Examples}

\begin{table}[tb]
    \small
    \begin{center}
    \begin{tabular}{ c | c  c  c  c }
    \hline \hline
    & $\alpha$ & FGSM & BIM(30,30) & Madry(30,30)  \\
    \hline
    \multirow{2}{*}{MNIST} & $e^{-3}$ & 96.37\% & 73.97\% & 54.75\% \\
    & $e^{-4}$ & 98.03\% & 19.91\% & 1.23\% \\
    \hline
    \multirow{2}{*}{FMNIST} & $e^{-3}$ & 83.59\% & 49.41\% & 32.31\% \\
    & $e^{-4}$ & 84.03\% & 33.82\% & 11.37\% \\
    \hline \hline
    \end{tabular}
    \end{center}
    \caption{Test Accuracy of the ATDA with Different $\alpha$}
    \vspace{-5mm}
    \label{table:atda-diff-lr}
\end{table}
Although we show that ATDA classifier has degenerated performance on BIM examples by analyzing feature space encoding, its numerical results, accuracy, in \cite{song2018improving} does not reflect the same finding. A possible reason is that the perturbation limit is too small compared with previous works such as \cite{madry2017towards}. Moreover, while experimenting with the source code, we locate another possible reason, the tuned learning rate $\alpha=e^{-3}$.

We evaluate ATDA classifiers with different values of $\alpha$ as shown in Table \ref{table:atda-diff-lr}. The results show that the test accuracy on both BIM(30,30) and Madry(30,30) examples significantly degenerates when we decrease $\alpha$. When $\alpha=e^{-4}$, the ATDA performs in a similar way as FGSM-Adv in \cite{tramer2017ensemble}. Actually, \cite{tramer2017ensemble} has shown that FGSM-Adv causes the trained classifier to overfit FGSM examples and be vulnerable to Iter-Exps. It is intuitive that optimizing with a smaller learning rate should converge to the same location if not a better location. Therefore, the degeneration in Table \ref{table:atda-diff-lr} means that the objective function in ATDA, combining FGSM-Adv and domain adaptation losses, is not appropriate for correct classification of Iter-Exps.

\begin{figure*}[tb]
\centering
\begin{minipage}[c]{.32\textwidth}
    \begin{minipage}[c]{\textwidth}
    \centering
        \includegraphics[width=\linewidth]{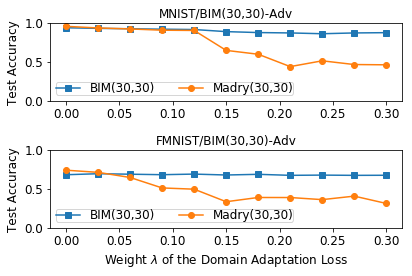}
    \end{minipage}
\captionsetup{width=.9\linewidth}
\caption{Evaluation of the Domain Adaptation Loss}
\label{fig:bim-adapt}
\end{minipage}
\begin{minipage}[c]{.315\textwidth}
    \begin{minipage}[c]{\textwidth}
    \centering
        \includegraphics[width=\linewidth]{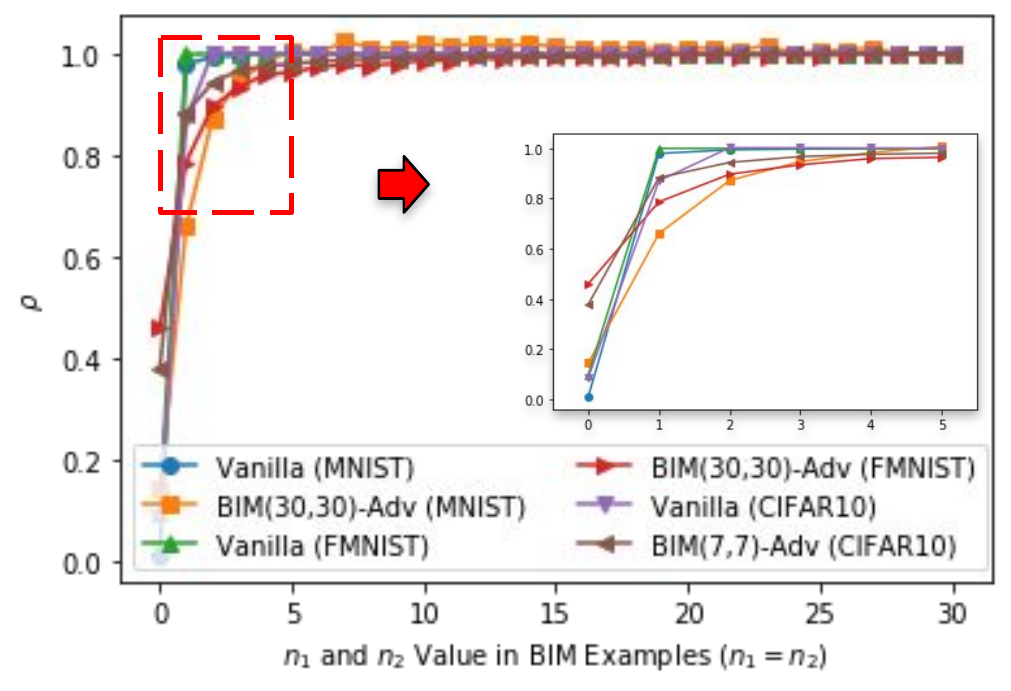}
    \end{minipage}
\captionsetup{width=.9\linewidth}
\caption{Experimental Results on Per Step Perturbation}
\label{fig:per-step}
\end{minipage}
\begin{minipage}[c]{.32\textwidth}
    \begin{minipage}[c]{\textwidth}
    \centering
        \includegraphics[width=\linewidth]{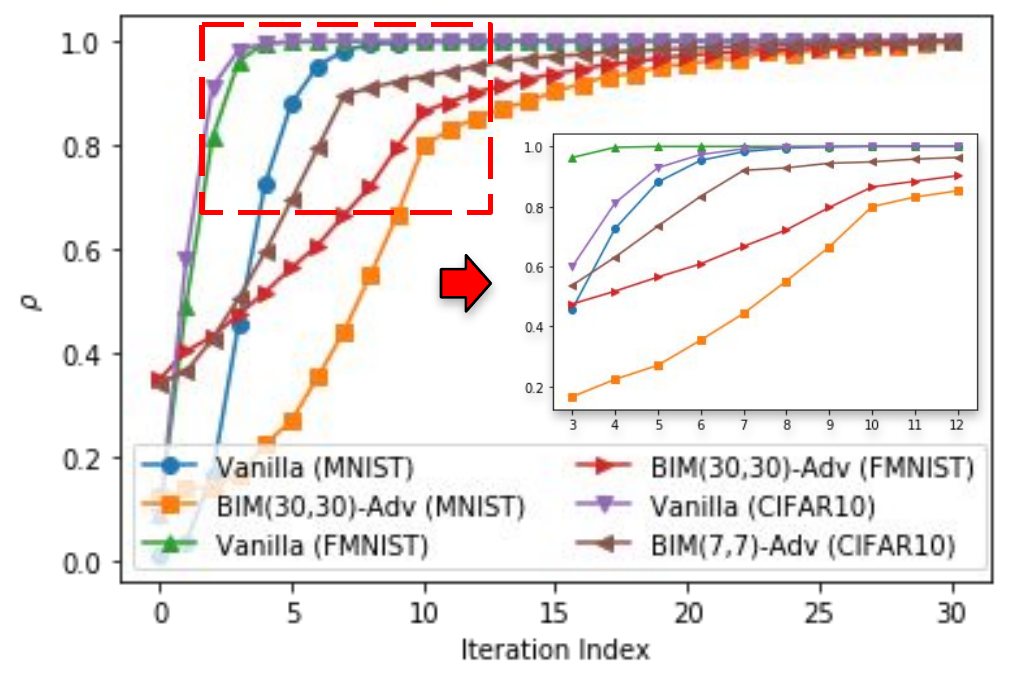}
    \end{minipage}
\captionsetup{width=.9\linewidth}
\caption{Experimental Results on Intermediate Examples}
\label{fig:intermediate}
\end{minipage}
\end{figure*}
\subsection{Vulnerability to Randomness}

Although previous subsections show that ATDA cannot defend Iter-Exps due to inappropriate objective function, the reason could solely be the usage of FGSM examples in the training. To pinpoint the effect of domain adaptation loss, we design another experiment which uses BIM examples instead of FGSM examples for training. We choose BIM(30,30)-Adv as the baseline and combine it with the domain adaptation loss proposed by ATDA. Our experiments are conducted on both MNIST and FMNIST datasets with BIM(30,30) and Madry(30,30) examples. Moreover, we assign different values to $\lambda$, a parameter that controls the weight of domain adaptation loss.

Figure \ref{fig:bim-adapt} presents the results of this experiment. When $\lambda = 0$, the classifier is solely trained with the cross-entropy loss. As $\lambda$ increases, the domain adaptation loss becomes more and more important in the total training loss. To our surprise, this experiment exposes another vulnerability of ATDA. Compared with cross-entropy loss, the domain adaptation loss does not make extra positive impact on the test accuracy. The test accuracy on BIM(30,30) examples remains unchanged or shows a small degeneration. Even worse, the domain adaptation loss hurts the test accuracy of the classifier on Madry(30,30) examples, especially when $\lambda \geq 0.15$. We think a reasonable explanation is that the randomness in Madry examples breaks the statistical assumption used in the domain adaptation loss.

\subsection{Summary}

In this section, we show that ATDA cannot defend Iter-Exps. Firstly, ATDA's feature space encoding shows that it miserably fails against Iter-Exps. Secondly, by changing the learning rate, we show that the objective function of the ATDA does not take Iter-Exps into consideration. Lastly, by combining domain adaptation loss with Iter-Adv, we show that relying on domain adaptation loss degenerates the accuracy of classifying Iter-Exps with randomness (Madry examples). \guanxiongA{Although the analysis is done in MNIST and FMNIST datasets, our further evaluation results on CIFAR10 dataset (Sec \ref{sec:evaluation}) also show that ATDA performs poorly in defending Iter-Exps.}


\section{Analysis of Iter-Adv Methods}\label{sec:iterative}

Compared with Single-Adv methods, Iter-Adv methods have significantly higher test accuracy. Therefore, the majority of adversarial training defenses, including the SOTA ones, focus on Iter-Adv methods. In spite of this, the domain is still not very well comprehended. In this section, we explore several fundamental questions regarding SOTA Iter-Adv methods through extensive set of experiments.

\subsection{Limit of Decreasing the Per Step Perturbation}

Based on the introduction of Iter-Exps in Section \ref{sec:introduction}, it is clear that the smaller per step perturbation applied, the better observation of NN's decision hyperplane and the stronger adversarial examples obtained. However, to select an appropriate per step perturbation, we believe that a quantitative analysis is needed.

\guanxiongA{
To do this analysis, we conduct experiments on MNIST, FMNIST, and CIFAR10 datasets. In each dataset, we train two different NN classifiers with the same structure and hyper-parameter settings, which include: (1) a Vanilla classifier trained on original examples only and (2) a BIM-Adv classifier \cite{kurakin2016adversarial2}. For each $n_{1} = n_{2}$ value in the range $\mathbb{Z}_{[0,30]}$, we generate BIM examples with fixed $\epsilon$ ($0.3$ in MNIST, $0.2$ in FMNIST, and $\frac{8}{255}$ in CIFAR10) and calculate the following ratio, $\rho$:
\begin{align}
    \rho={\frac{\text{error rate under current value of $n_{1}$ and $n_{2}$}}{\text{error rate under the maximum value of $n_{1}$ and $n_{2}$}}} \label{eq:percentage}
\end{align}
To understand $\rho$, assume that its value is $1$ when, for example, $n_{1} = n_{2} = 10$. And, the maximum value of $n_{1}$ and $n_{2}$ is $30$ (e.g. with MNIST dataset). This means that BIM examples generated with value $n_{1} = n_{2} = 10$ can be as successful as those generated with value $n_{1} = n_{2} = 30$ in misleading the classifier.
}

From the results in Figure \ref{fig:per-step}, it is clear that $\rho$ converges fast and saturates when the value of $n_{1} = n_{2}$ is around 5 in all six lines. For the Vanilla classifiers, this phenomenon is not surprising since they have no defence at all and most of the adversarial examples can fool them. However, we see a similar trend from the BIM-Adv classifiers which are well trained to defend adversarial examples. The insight we draw from this experiment is that increasing the value of $n_{1} = n_{2}$ over a certain limit provides only marginal help in finding stronger adversarial examples. In other words, training a classifier by Iter-Adv method with small $n_{1} = n_{2}$ values (around 5 in this experiment) is as efficient as training the classifier by Iter-Adv with large $n_{1} = n_{2}$ values (30 in this experiment).

Given the fact that adversarial training uses adversarial examples to find blind spots of the under-trained classifier and retrain it, these results show that \textit{decreasing the per step perturbation of Iter-Exps used during Iter-Adv beyond a certain limit only marginally benefits the defense.}

We think the saturation of per step perturbation exists because the loss structure used in projected gradient descent to search Iter-Exps is shown to be highly tractable \cite{madry2017towards}. This important finding indicates that defenses could use smaller values of $n_{1} = n_{2}$ without sacrificing the quality of the defense. Although the resulting defense is still within the Iter-Adv category, it consumes less time and computations in preparing adversarial examples. We will utilize this observation in combination with the following others to develop an efficient Single-Adv method.

\subsection{Training with Intermediate Adversarial Examples}

As shown in Section \ref{sec:background} Eq. \ref{eq:iter-exps}, Iter-Adv methods usually use final adversarial examples ($x_{n_{2}}$) to build the defense since it is much stronger than intermediate versions ($x_{i}, ~ \forall i < n_{2}$). In this section, we explore whether those intermediate examples can be utilized for training while being generated, instead of sitting idle and waiting for the final versions.

\guanxiongA{
To do so, we conduct another set of experiments on MNIST, FMNIST and CIFAR10 datasets. In these experiments, we use the same NN classifiers, measure the same ratio ($\rho$ as defined in Eq. \ref{eq:percentage}), and maintain the same perturbation limit used in the preceding subsection. The only difference is that we here fix the value of $n_{1}$ (10 for MNIST and FMNIST, 7 for CIFAR10) and $n_{2}$ (30 for all) in BIM examples. Values along the X-axis in Figure \ref{fig:intermediate} correspond to different iterations during the generation of BIM examples. For example in MNIST, an X-axis value of zero corresponds to the original examples, a value of $0<i<30$ represents the corresponding intermediate examples after $i$ iterations, and a value of 30 corresponds to the final versions, BIM(10,30), of the adversarial examples.
}

\guanxiongA{
The results show that $\rho$, under all the scenarios, is monotonically increasing with the number of iterations. For all three Vanilla classifiers which have no defense against adversarial examples, $\rho$ saturates quickly after around 5 iterations. Although $\rho$ for BIM-Adv classifiers does not saturate as quick as that for Vanilla classifiers, it increases almost exponentially before reaches around $0.8$ (MNIST and FMNIST) or $0.9$ (CIFAR10). In the zoom-in view of Figure \ref{fig:intermediate}, we can clearly see these turning points. One interesting observation is that the turning point in BIM-Adv classifier corresponds the selected value of $n_{1}$ (10 or 7).
}

Our insights from this experiment is that: \textit{a large portion of intermediate examples of Iter-Exps are good quality adversarial examples since they effectively reveal the classifier's blind spot, and hence can be used to build high quality defenses.} In other words, we can utilize the intermediate examples during the preparation of Iter-Exps to continuously enhance the model instead of waiting for the final examples while training Iter-Adv defenses. In Section \ref{sec:methodology}, we build on this finding to expand the generation process of Iter-Exps and end up with a Single-Adv method that performs very close to Iter-Adv.

\subsection{Summary}

\guanxiongA{In this section, we identify two experimental properties of Iter-Adv: (1) It is recommended to use large per step perturbation, i.e., small $n_{1}$ and (2) intermediate examples can be used to speed up adversarial training with a very small degeneration in Iter-Exps quality.}

\begin{figure*}[tb]
\centering
\begin{minipage}[c]{.32\textwidth}
    \begin{minipage}[c]{\textwidth}
    \centering
        \includegraphics[width=\linewidth]{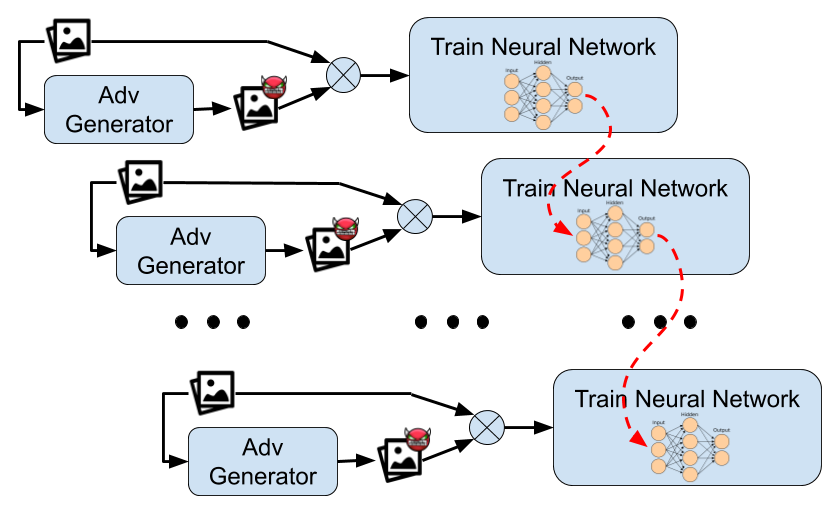}
    \end{minipage}
\captionsetup{width=.9\linewidth}
\caption{Training with Single-Exps or Iter-Exps}
\label{fig:adv-algorithm}
\end{minipage}
\begin{minipage}[c]{.32\textwidth}
    \begin{minipage}[c]{\textwidth}
    \centering
        \includegraphics[width=\linewidth]{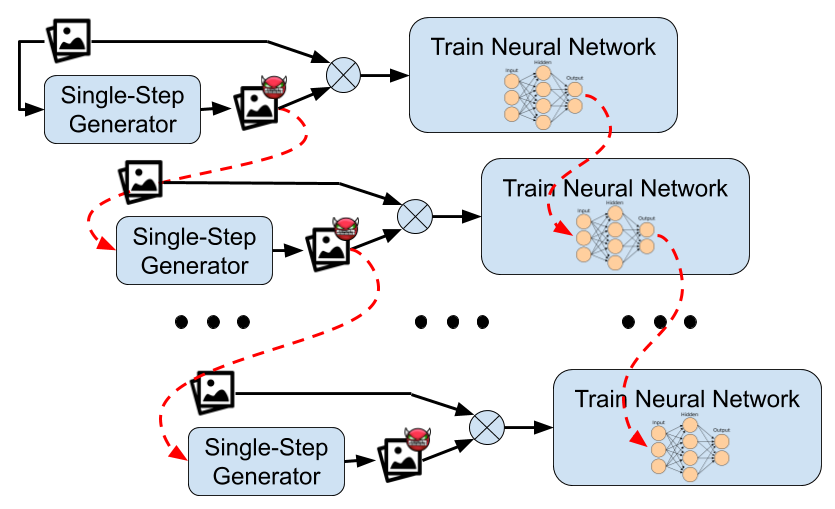}
    \end{minipage}
\captionsetup{width=.9\linewidth}
\caption{Training with Single-Step Epoch-Iterative Examples}
\label{fig:si-algorithm}
\end{minipage}
\begin{minipage}[c]{.30\textwidth}
    \begin{minipage}[c]{\textwidth}
    \centering
        \includegraphics[width=\linewidth]{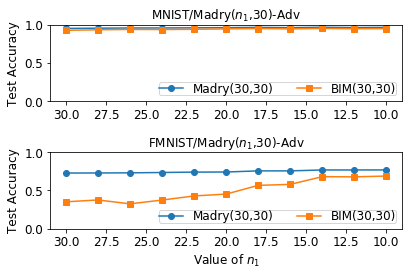}
    \end{minipage}
\captionsetup{width=.9\linewidth}
\caption{Madry-Adv under Different Hyper-Parameter Settings}
\label{fig:over-perturbation-madry}
\end{minipage}
\end{figure*}

\section{Single-Step Epoch-Iterative Method}\label{sec:methodology}

In previous sections, we conduct experiments and evaluate both Single-Adv and Iter-Adv methods in details. The insights from the results help us enhance our understanding of adversarial training and its underlying fundamental concepts. In this section, we propose a new Single-Adv method which we call Single-Step Epoch-Iterative Method.

\subsection{Motivation and Design}

In Figure \ref{fig:adv-algorithm}, we review the process of adversarial training. Each row in the figure represents a training epoch and the solid black lines represent the data flow (training examples). The dashed red lines across different rows correspond to knowledge flow (classifier's weights) from one epoch to the next. As shown in the figure, the original examples are fed into the generator of adversarial examples, which could be single-step or iterative. Then, the original and adversarial examples are used to train the classifier. The training process consists of several training epochs and the weights of the classifier are carried out from one training epoch to the next.

\guanxiongA{Inspired by the empirical findings drawn from our previous experiments (Section \ref{sec:iterative}), we propose the following modifications to enhance the Single-Adv defense process. Similar to other Single-Adv methods, our method also uses the single-step generator to reduce computation overhead in each epoch. Recall that a classifier which is trained with Single-Adv fails to defend Iter-Exps, therefore, we use consecutive training epochs to mimic the generation of Iter-Exps.}

\guanxiongA{Starting from the second training epoch, we reuse the output of the generator from the previous epoch as input to the generator of the current epoch, instead of using the original image. As a result, the classifier can be seen as trained with intermediate examples in the first $(n_{2}-1)$ training epochs. In each training epoch, our method uses a relatively large per step perturbation (i.e., small $n_{1}$) instead of total perturbation (i.e., $n_{1} = 1$). It helps our method to avoid repeatedly generating Single-Exps for training. On the other hand, a large per step perturbation ensures the adversarial examples can quickly reach their maximum perturbation. Therefore, it can mitigate the degeneration caused by training with weak intermediate examples in the first few training epochs. Lastly, after repeating aforementioned steps for $n_{2}$ epochs, the generator switches to select original examples as inputs which means the iteration over consecutive epochs is reset.}

\guanxiongA{
From a high level point of view, we flatten the iteration of generating Iter-Exps into training epochs. Within each iteration of $n_{2}$ consecutive training epochs, the mathematical formulation of generating adversarial examples is as follows.
\begin{align}
    & \delta_{i+1} = clip_{[-\epsilon,\epsilon]} [\frac{\epsilon}{n_{1}} \times sign[\nabla_{\tilde{x}_{i}} L(\tilde{x}_{i}, y, \theta)]]\\
    & \tilde{x}_{i+1} = clip_{[0,1]} [\tilde{x}_{i} + \delta_{i+1}]\\ 
    & \tilde{x}_{0} = \hat{x} \hspace{5mm} i \in [0,n_{2}]
\end{align}
}
\guanxiongA{
Here, $i$ represents the index of iteration over training epochs. Similar as traditional adversarial training method, we also present the process of SIM-Adv as a flow chart in Figure \ref{fig:si-algorithm}.
}

\subsection{Applying Over-Perturbation}

\guanxiongA{
In the previous subsection, we present the core design of using Single-Exps to mimic Iter-Exps. At the same time, we also mention the potential disadvantage of this design. In the majority of training epochs, our method uses the intermediate examples. Recall the experiment results in Figure \ref{fig:intermediate}, these intermediate examples are less serious than the corresponding Iter-Exps. Based on our experimental results, the classifier directly trained with the SIM examples can defend against adversarial examples but performs worse than that trained with Iter-Exps.
}

\guanxiongA{
To further mitigate the gap in performance, we now introduce a heuristic modification of the hyper-parameter setting which we call it \textbf{over-perturbation}. \textit{For the first time, we define two different hyper-parameter settings in adversarial training methods. We define that the setting is over-perturbation when $n_{1} < n_{2}$, otherwise is under-perturbation.} This modification is based on our empirical results. As aforementioned, the intermediate examples are less serious than final Iter-Exps. However, the zoom-in view in Figure \ref{fig:intermediate} shows the existence of turning point in the iteration index and its connection to the value of $n_{1}$. Before the turning point, the success rate to mislead classifiers by intermediate examples is much lower than that by Iter-Exps but increases exponentially, and vice versa.
}

\guanxiongA{
By applying over-perturbation, we actually ensure that our method trains the classifier with strong adversarial examples in most of the training epochs. Assume we run 20 epochs of training with two settings (1) $n_{1} = n_{2} = 10$ and (2) $2n_{1} = n_{2} = 20$. Under the first setting, the classifier is trained with intermediate examples before the turning point in both 1$^{st}$ to 9$^{th}$ and 11$^{th}$ to 19$^{th}$ epochs. While, under the second setting, the intermediate examples used between $10^{th}$ and $19^{th}$ epoch are after the turning point. Overall, the second setting spends more epochs in training the classifier with strong adversarial examples and the trained classifier performs better on defending adversarial examples.
}

\guanxiongA{
Actually, the over-perturbation setting has been used in previous research works, intentionally or unintentionally. For example in \cite{madry2017towards}, the hyper-parameter setting of all Madry-Adv methods are over-perturbation. For curiosity, we try to change the hyper-parameter setting from under to over-perturbation and record the performance. In both MNIST and FMNIST datasets, we fix the $n_{2}$ to 30 and iterativelly reduce the value of $n_{1}$ from 30 to 10. In each setting, we measure classifier's test accuracy on both BIM and Madry examples. These results are presented in Figure \ref{fig:over-perturbation-madry}. To our surprise, we found that the Madry-Adv with under-perturbation may train a classifier performs significantly worse than that trained with over-perturbation (in FMNIST).
}

\guanxiongA{
For this phenomenon, we believe that it is related to the random initialization in Madry-Adv. In some situations, the random initialization may add unnecessary perturbation to the image and such perturbation could degenerate the performance of trained classifier. Under such situations, the over-perturbation makes it possible to eliminate unnecessary perturbation which enhances the performance of the trained classifier. More detailed analysis of combining over-perturbation and Madry-Adv is out of the scope of this work and we keep it as our future work.
}

\begin{figure*}[tb]
\centering
\begin{minipage}[c]{.35\textwidth}
    \begin{minipage}[c]{\textwidth}
    \centering
        \includegraphics[width=\linewidth]{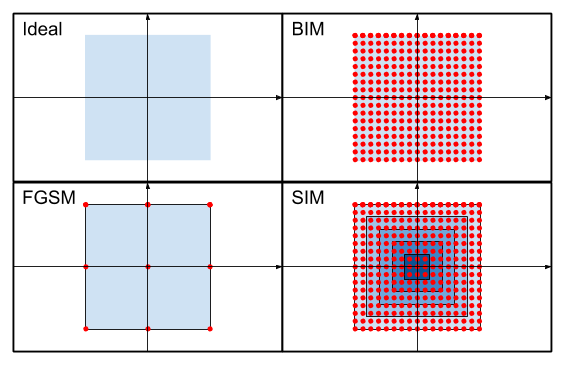}
    \end{minipage}
\caption{Searching Space of Different Adversarial Examples}
\label{fig:toy-example}
\label{fig:adv-algorithm}
\end{minipage}
\begin{minipage}[c]{.55\textwidth}
    \begin{minipage}[c]{\textwidth}
    \centering
        \includegraphics[width=\linewidth]{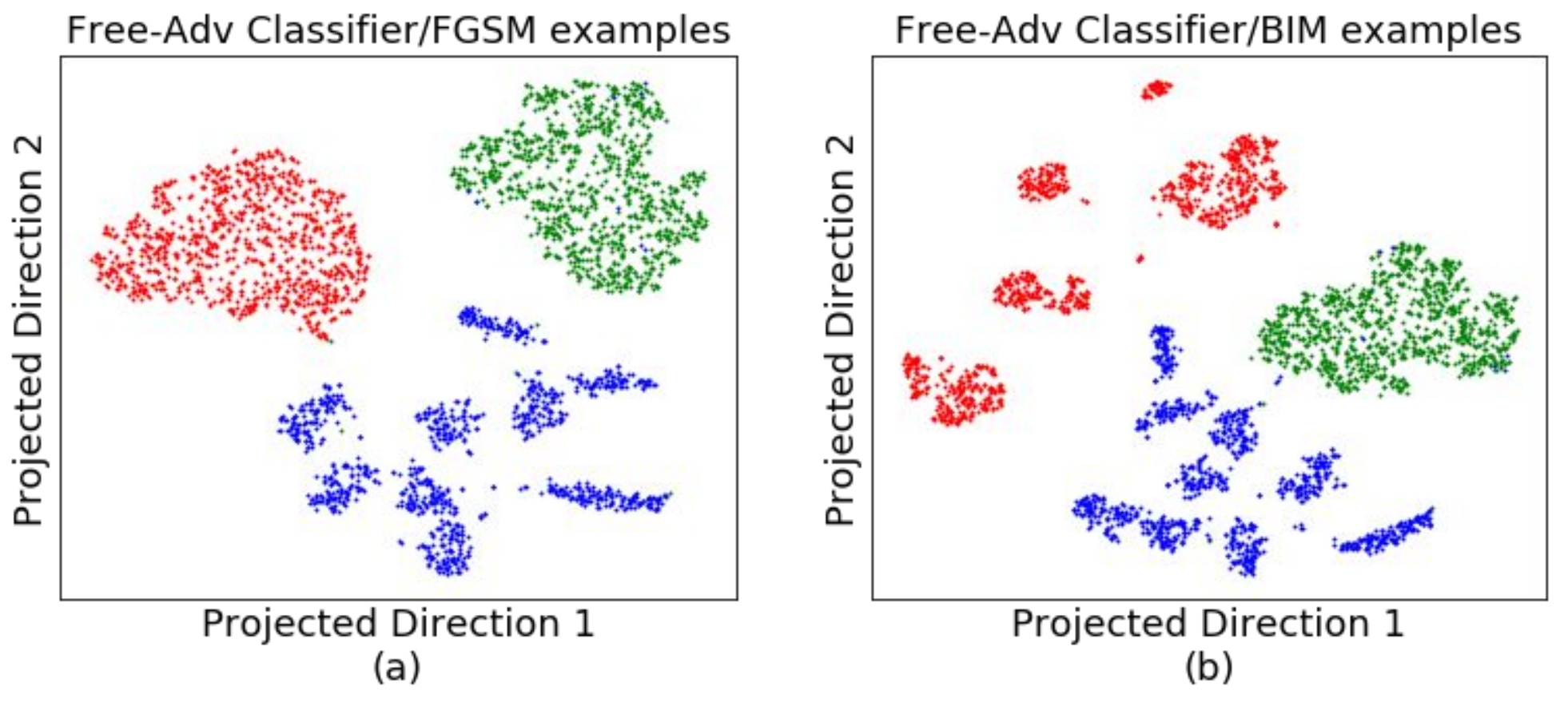}
    \end{minipage}
\captionsetup{width=.9\linewidth}
\caption{Feature Space Encoding of Free-Adv Classifier}
\label{fig:feature-free-adv}
\end{minipage}
\end{figure*}
\subsection{Searching Space for Adversarial Examples} 

\guanxiong{
In order to intuitively show the effectiveness of our proposed method, we prepare a toy example to compare the search space of adversarial examples. Without loss of generality, we assume that the data in this toy example is in a 2 dimensional space. Therefore, we could easily present the search space of adversarial examples. In Figure \ref{fig:toy-example}, we choose to present the search space of optimal adversarial examples as well as BIM, FGSM and SIM examples.
}

\guanxiong{
As we see from the top-left corner, the search space of optimal adversarial examples is the blue square which represents the entire norm ball. However, this search space corresponds to the exhaustive search which cannot be achieved. Among others, the BIM examples are the best mimic of the optimal adversarial examples. Since it separate the total perturbation into multiple steps and iteratively apply small perturbations, the search space of BIM examples can be represented by the mesh of red dots. The density of dots is related to the size of per step perturbation. Compared with BIM examples, the search space of FGSM examples are significantly limited. Since the total perturbation is applied all at once, the potential locations (red dots) of FGSM examples can only cover corners and some surface of the norm ball, while the entire inner space between origin and perturbation boundary is unreachable.
}

\guanxiong{
When we focus on the SIM examples, its search space could be represented by a mesh of red dots with dynamically changing size. With a fixed density of dots, the size of the mesh increases from the smallest one (just around the origin) to the largest one (same size as the entire norm ball) epoch-by-epoch. Although the searching space of SIM examples is smaller than that of BIM examples at the beginning, small $n_{1}$ value and over-perturbation setting ensure that most of epochs are searching adversarial examples in the entire norm ball. Moreover, the analysis of Iter-Exps shows that a relatively lower density of dots does not significantly affect the searching of adversarial examples. Last but not the least, SIM examples are Single-Exps which consumes less computation overhead compared to the Iter-Exps of BIM.
}

\subsection{Comparing with the Free-Adv}

\guanxiong{
During the preparation of this work, we noticed that there is a related work, denoted as \textbf{Free-Adv}, that is uploaded to the arXiv \cite{shafahi2019adversarial}. This work shares a similar design to our paper in generating adversarial examples. However, our proposed work is different from it and performs better. Until writing this paper, this work is still an arXiv paper and has not been accepted as a conference or journal submission. Therefore, we did not list it as the SOTA Single-Adv method due to the lack of peer review. However, for the completeness of our work, we still reimplement, evaluate and compare it with our proposed method.
}

During the generation process of adversarial examples, both the Free-Adv and our proposed method are reusing the gradient information. In the Free-Adv, the training images in each mini-batch are Single-Exps. In order to mitigate the issue of training with Single-Exps, the authors try to replay each mini-batch multiple times (denoted as $m$) and reuse the adversarial examples from previous replay as the inputs. Finally, the total training epoch of the Free-Adv will be decreased by the factor of the total replay iterations. Although the Free-Adv looks similar to our proposed method, there are two differences which can distinguish it from our work.

\guanxiong{
First of all, the proposed algorithm in the Free-Adv replays the images in the mini-batch immediately which means the classifier is repeatedly trained with the same mini-batch several times. As the authors pointed out in \cite{shafahi2019adversarial}, such replay could cause the “catastrophic forgetting” problem. If the information in the training data is unbalanced, the classifier could become overfitting through being repeatedly trained with a certain category of training data. On the contrary, our proposed method utilizes gradient information across different training epochs. In other words, our method does not change the training order which could be specially designed. Therefore, our method does not have the risk of causing the “catastrophic forgetting” problem \cite{shafahi2019adversarial}.
}

\guanxiong{
Secondly, in our work adversarial perturbation is applied in a different way compared with Free-Adv. Based on Algorithm 2 in \cite{shafahi2019adversarial} and the official implementation on GitHub, it is clear that the Free-Adv utilizes the total perturbation during each replay. From Figure \ref{fig:toy-example}, we show that using total perturbation only (e.g. FGSM examples) has limited sampling locations which can only cover the corners and parts of the surface of a $l_{\infty}$ ball. Although the Free-Adv repeatedly replays the mini-batch several times, its sampling locations will not change unless most of the pixel values are close to the clipping boundary. Although people may think that it is a problem about the hyper-parameter tuning, we argue that the design of applying adversarial perturbation should be based on the empirical analysis of iterative adversarial examples proposed in Section \ref{sec:iterative}.
}

\guanxiong{
In order to support our hypothesis, we repeat the experiment in Section \ref{sec:atda} to represent the feature space encoding of the classifier trained with the Free-Adv. The results are presented in Figure \ref{fig:feature-free-adv} and the legends are the same as those in Figure \ref{fig:atda-toy}. When the FGSM examples are used as adversarial inputs, the classifier could group them together. However, these adversarial inputs and their corresponding targets belong to two separated groups which means the Free-Adv performs worse than the SIM-Adv. If the adversarial inputs are BIM examples, the trained classifier is fooled since the red dots are separated into several small groups. Overall, the presented results in Figure \ref{fig:feature-free-adv} reflect that the Free-Adv has a similar issue as the ATDA and performs even worse than ATDA in MNIST dataset. This issue is also identified during our evaluation of the Free-Adv with the details presented in the next section.
}

\begin{table}[tb]
    \small
    \begin{center}
    \begin{tabular}{ c | c  c  c }
    \hline \hline
    & MNIST & FMNIST & CIFAR10 \\
    \hline
    $\epsilon$ & $0.3$ & $0.2$ & $\frac{8}{255}$ \\
    $\tilde{x}$ & \multicolumn{3}{c}{FGSM, BIM and Madry examples} \\
    Single-Adv & \multicolumn{3}{c}{Free-Adv and ATDA} \\
    Iter-Adv & \multicolumn{3}{c}{BIM-Adv and Madry-Adv} \\
    Network Structure & \multicolumn{2}{c}{LeNet} & ResNet \\
    Metric & \multicolumn{3}{c}{test accuracy and total training time} \\
    Platform & \multicolumn{3}{c}{CleverHans} \\
    \hline \hline
    \end{tabular}
    \end{center}
    \caption{Evaluation Parameter Setting}
    \vspace{-3mm}
    \label{table:eval-setting}
\end{table}
\begin{table*}[tb]
    \small
    \begin{center}
    \begin{tabular}{ c  c || c | c  c  c | c  c }
    \hline \hline
    & & Vanilla & Free-Adv ($m=10$) & ATDA & SIM(10,20)-Adv & BIM(10,30)-Adv & Madry(10,30)-Adv \\
    \hline
    \multirow{4}{*}{\begin{sideways}MNIST\end{sideways}} & Original & 98.84\% & 98.72\% & 97.55\% & 99.00\% & 99.01\% & 99.01\% \\
    & FGSM & 4.46\% & 95.94\% & 96.37\% & 96.57\% & 96.56\% & 97.03\% \\
    & BIM(10,40) & 0.94\% & 1.56\% & 42.34\% & 92.55\% & 93.83\% & 94.04\% \\
    & Madry(10,40) & 0.87\% & 0.70\% & 33.14\% & 92.89\% & 94.15\% & 94.29\% \\
    \hline
    \multirow{4}{*}{\begin{sideways}FMNIST\end{sideways}} & Original & 91.64\% & 89.45\% & 84.64\% & 88.96\% & 86.19\% & 87.14\% \\
    & FGSM & 6.07\% & 85.69\% & 83.59\% & 79.78\% & 78.34\% & 75.82\% \\
    & BIM(10,40) & 5.96\% & 7.39\% & 25.81\% & 65.93\% & 69.71\% & 64.59\% \\
    & Madry(10,40) & 4.55\% & 3.99\% & 23.79\% & 66.88\% & 70.58\% & 69.72\% \\
    \hline
    & & Vanilla & Free-Adv ($m=8$) & ATDA & SIM(2,10)-Adv & BIM(4,7)-Adv & Madry(4,7)-Adv \\
    \hline
    \multirow{4}{*}{\begin{sideways}CIFAR10\end{sideways}} & Original & 91.74\% & 73.96\% & 89.11\% & 77.21\% & 80.83\% & 81.08\% \\
    & FGSM & 17.89\% & 48.96\% & 65.77\% & 54.12\% & 56.32\% & 56.08\% \\
    & BIM(4,20) & 5.56\% & 38.68\% & 8.02\% & 43.69\% & 46.77\% & 46.73\% \\
    & Madry(4,20) & 4.82\% & 38.63\% & 8.32\% & 43.78\% & 46.80\% & 46.74\% \\
    \hline \hline
    \end{tabular}
    \end{center}
    \caption{Test Accuracy}
    \label{table:summary-test-accuracy}
\end{table*}
%

\section{Evaluation}\label{sec:evaluation}

In this section, we first summarize the evaluation settings. Then, we present, analyze and compare the evaluation results of our proposed adversarial training method, SIM-Adv, with other defense methods.

\subsection{Evaluation Setting}


\guanxiong{
In this evaluation, we select three different datasets: MNIST, FMNIST and CIFAR10. For the MNIST and FMNIST datasets, we select the LeNet \cite{lecun1998gradient} as network structure, while, the ResNet structure \cite{he2016deep} is used in the CIFAR10 dataset. Within each dataset, we evaluate the performance (classification accuracy) of the trained classifier against both original and different types of adversarial examples.
}

As mentioned in Section \ref{sec:background}, all the adversarial examples throughout this work are $l_{\infty}$ white-box untargeted adversarial examples in the image classification domain. \guanxiongA{For the method of generating adversarial examples, our selection includes FGSM, BIM and Madry which reflects the performance on both Single-Exps (FGSM examples) and Iter-Exps (BIM and Madry examples).} The total perturbation limits are 0.3 in MNIST, 0.2 in FMNIST and $\frac{8}{255}$ in CIFAR10 which is used in the \cite{madry2017towards}. To make the evaluation more convincing, we select larger $n_{2}$ value in both BIM and Madry examples to make them stronger than adversarial examples used during training.

As a baseline, we present the evaluation results of the vanilla classifier, a one with no defense against adversarial examples. To better evaluate our proposed method, we compare not only with Single-Adv methods (ATDA and Free-Adv), but also with Iter-Adv ones (BIM-Adv and Madry-Adv). \guanxiongA{In the evaluation, we skip adversarially trained models with FGSM or R+FGSM examples. Although FGSM-Adv and R+FGSM-Adv are single-step version of BIM-Adv and Madry-Adv, previous studies show that they are failed to defend Iter-Exps \cite{kurakin2016adversarial1}\cite{tramer2017ensemble}. In other words, directly setting $n_{1}=n_{2}=1$ in BIM-Adv or Madry-Adv cannot train a classifier that makes correct predictions on Iter-Exps. Instead, we present the ATDA and Free-Adv as representatives of Single-Adv.}

\guanxiongA{For these selected adversarial training methods, we follow the original setting of hyper-parameters presented in their work and fine-tune the setting to present the best performance. For our proposed method, we tune its hyper-parameters as follows. Firstly, its $n_{1}$ can be select as the same or one half of $n_{1}$ value in corresponding BIM-Adv method. Secondly, the $n_{2}$ value is selected among 2 to 5 times of the $n_{1}$ value due to the over-perturbation. Among these combinations, we present its best performance. The specific hyper-parameter values of all methods are presented along with evaluation results.}

\guanxiong{
During the evaluation, we compare different trained classifiers in terms of test accuracy which is defined as follows:
\begin{align}
    \text{test accuracy} \equiv \frac{\text{\# of correctedly classified inputs}}{\text{\# of total inputs}}
\end{align}
Here, the calculation of test accuracy is based on a single category of inputs (e.g. original examples or FGSM examples). Moreover, we also measure the total training time consumed by training classifiers with different methods. To ensure the quality and reproducibility, adversarial examples used in both training and evaluation are based on the well-known standard python platform, CleverHans \cite{papernot2018cleverhans}. A summary of these evaluation settings is also presented in Table \ref{table:eval-setting}.
}

\begin{table}[tb]
    \small
    \begin{center}
    \begin{tabular}{ c | c  c  c  c }
    \hline \hline
    & $\alpha$ & FGSM & BIM(10,40) & Madry(10,40)  \\
    \hline
    \multirow{2}{*}{MNIST} & $e^{-3}$ & 93.87\% & 89.93\% & 90.11\% \\
    & $e^{-4}$ & 96.57\% & 92.55\% & 92.89\% \\
    \hline
    \multirow{2}{*}{FMNIST} & $e^{-3}$ & 74.68\% & 61.07\% & 61.46\% \\
    & $e^{-4}$ & 79.78\% & 65.93\% & 66.88\% \\
    \hline \hline
    \end{tabular}
    \end{center}
    \caption{Test Accuracy of SIM-Adv with Different $\alpha$}
    \vspace{-3mm}
    \label{table:sim-double}
\end{table}
%
\subsection{Test Accuracy}

\guanxiong{
We first evaluate the Free-Adv method. As Table \ref{table:summary-test-accuracy} shows, the Free-Adv classifier can defend both Single-Exps and Iter-Exps in the CIFAR10 dataset. However, its Iter-Exps accuracy degenerates significantly with MNIST and FMNIST datasets. We think this phenomenon is related to the second issue of the Free-Adv that is mentioned in Section \ref{sec:methodology}. The limited searching space makes the classifier trained with the Free-Adv unable to defend the Iter-Exps. In the CIFAR10 dataset, training with the Free-Adv enhances the defence against Iter-Exps since the per step perturbation is enlarged from $\frac{1}{10} \epsilon$ to $\frac{1}{4} \epsilon$. As a result, the issue of limited searching space is mitigated. Even in the CIFAR10 dataset, the test accuracy of the classifier trained with the Free-Adv is still lower than that of our classifier (SIM-Adv), because SIM-Adv can tune the hyper-parameters (i.e. $n_{1}$) for more appropriate per step perturbations.
}

As shown in Table \ref{table:summary-test-accuracy}, the performance of the SOTA Single-Adv (ATDA) is significantly worse than that in \cite{song2018improving}. The reason is that the perturbation limit is too small in the original work. For example, $\epsilon$ in the original evaluation is $\frac{4}{255}$ instead of $\frac{8}{255}$ on CIFAR10 dataset. As a result, Iter-Exps are similar to Single-Exps and even FGSM-Adv achieves over 49\% test accuracy on Madry examples. Therefore, we think the original evaluation of ATDA is misleading. Our exclusive experiments in Section \ref{sec:atda} point that ATDA actually fails to defend Iter-Exps due to the use of: (1) FGSM examples which are less representative, and (2) domain adaptation loss that is vulnerable to randomness.

Compared with the Free-Adv and the ATDA methods, the evaluation results show that adversarial training with SIM examples (SIM-Adv) achieves better and more stable performance. In all the three datasets, the classifier trained with the SIM-Adv can defend both Single-Exps and Iter-Exps while maintaining a reasonable test accuracy of original examples. More importantly, the SIM-Adv significantly enhances the test accuracy on Iter-Exps over the ATDA and the Free-Adv. To the best of our knowledge, this is the first Single-Adv method which can efficiently defend Iter-Exps under the white-box adversary model. To double check, we also test the SIM-Adv with the same evaluation on the ATDA work in Section \ref{sec:atda}. The results show that the SIM-Adv can defend both Single-Exps and Iter-Exps under different values of $\alpha$ and achieve better performance when $\alpha$ is lower. The evaluation results are presented in Table \ref{table:sim-double}.

\guanxiong{
To make our evaluation results more convincing, we also compare the SIM-Adv with SOTA Iter-Adv methods which include both BIM-Adv and Madry-Adv. The overall results show that the SIM-Adv has a competitive performance as the BIM-Adv and Madry-Adv. Although the classifier trained with the SIM-Adv has a degeneration in terms of test accuracy, we think the less than 4\% decrease is a reasonable trade-off given that the SIM-Adv trains the classifier with weak adversarial examples in some of its training epochs to save training overhead. 
}
Moreover, we could tune the hyper-parameter of the SIM-Adv to achieve better trade-off between test accuracy and training cost. For example, the SIM-Adv could perform a two-step iteration in each training epoch instead of single-step. We leave the analysis of this for future works.

\begin{table}[tb]
    \small
    \begin{center}
    \begin{tabular}{ c | c  c  c }
    \hline \hline
    & MNIST & FMNIST & CIFAR10 \\
    \hline
    Free-Adv & 234.75 & 308.5 & 25923.5 \\
    ATDA & 319.56 & 422.4 & 33011.6 \\
    SIM-Adv & 293.22 & 391.2 & 26692.4 \\
    BIM-Adv & 866.76 & 1159.2 & 111372.6 \\
    \hline \hline
    \end{tabular}
    \end{center}
    \caption{Total Training Time in Seconds}
    \vspace{-3mm}
    \label{table:summary-time}
\end{table}
\subsection{Training Time}

To fairly evaluate the total training time, we select four different defense methods: the Free-Adv, the ATDA, the SIM-Adv and the BIM-Adv. We do not present the Madry-Adv since it has similar training time as that of BIM-Adv. All of our results are measured on a Dell Workstation with a NVIDIA RTX-2070 GPU.

The evaluation results in Table \ref{table:summary-time} clearly show that our SIM-Adv significantly reduces the total training time compared with the BIM-Adv. The SIM-Adv saves more than 60\% on both MNIST and FMNIST datasets and more than 75\% on CIFAR10 dataset in terms of total training time. Even compared with the ATDA, our SIM-Adv can still save at least 7\% of total training time since the domain adaptation loss requires additional computation. When comparing with the Free-Adv, our proposed SIM-Adv consumes more training time since it has to save and restore the gradient information across training epochs.


\section{Conclusion}\label{sec:conclusion}

In this work, we first show through thorough empirical analysis that the SOTA Single-Adv method (ATDA) is not well evaluated and fails to defend against Iter-Exps. We also provide thorough empirical analysis of SOTA Iter-Adv defense methods and draw insights that can help enhance future Iter-Adv defense methods. In particular, we show that (1) using larger per step perturbation does not hurt the performance of Iter-Adv, (2) the intermediate examples in preparing Iter-Exps reveal the majority of classifier's blind spots. Finally, we propose a novel Single-Adv defense method, SIM-Adv, and highlight its advantages over a recent related work, Free-Adv. We show through comparative experiments that SIM-Adv is the first Single-Adv defense method that can efficiently defend against both Single-Exps and Iter-Exps with much lower training overhead compared to the SOTA Iter-Adv counterparts. 

\bibliographystyle{ACM-Reference-Format}
\bibliography{ref.bib}

\end{document}